\pdfoutput=1

\documentclass[11pt]{article}

\usepackage{acl}
\usepackage{hyperref}
\usepackage{times}
\usepackage{latexsym}
\usepackage{tikz}
\usepackage{amssymb}
\usepackage{amsmath}
\usepackage{inconsolata}
\usepackage{microtype}
\usepackage{CJKutf8}
\usepackage{subfigure}
\usepackage{authblk}
\usepackage{caption}
\usepackage{algorithm}
\usepackage{algorithmic}
\usepackage{booktabs}
\usepackage{makecell}
\usepackage[T1]{fontenc}
\usepackage[utf8]{inputenc}

\title{DoRA: Enhancing Parameter-Efficient Fine-Tuning with Dynamic Rank Distribution}

\author[1,2]{Yulong Mao\thanks{Equal contributions.}}
\author[1,2]{Kaiyu Huang\protect\footnotemark[1]}
\author[1,2]{Changhao Guan}
\author[1,2]{\\Ganglin Bao}
\author[3]{Fengran Mo}
\author[1,2]{Jinan Xu\thanks{Corresponding author.}}

\affil[1]{Beijing Key Lab of Traffic Data Analysis and Mining, Beijing, China}
\affil[2]{Beijing Jiaotong University, Beijing, China}
\affil[3]{Université de Montréal, Montréal, Canada}

\affil[ ]{\{maoyulong, kyhuang, jaxu\}@bjtu.edu.cn}

\begin{document}
\maketitle

\begin{abstract}
Fine-tuning large-scale pre-trained models is inherently a resource-intensive task. While it can enhance the capabilities of the model, it also incurs substantial computational costs, posing challenges to the practical application of downstream tasks. Existing parameter-efficient fine-tuning (PEFT) methods such as Low-Rank Adaptation (LoRA) rely on a bypass framework that ignores the differential parameter budget requirements across weight matrices, which may lead to suboptimal fine-tuning outcomes. To address this issue, we introduce the Dynamic Low-Rank Adaptation (DoRA) method. DoRA decomposes high-rank LoRA layers into structured single-rank components, allowing for dynamic pruning of parameter budget based on their importance to specific tasks during training, which makes the most of the limited parameter budget. Experimental results demonstrate that DoRA can achieve competitive performance compared with LoRA and full model fine-tuning, and outperform various strong baselines with the same storage parameter budget. Our code is available at \url{https://github.com/MIkumikumi0116/DoRA}
\end{abstract}

\section{Introduction}
Pre-trained Language Models (PLMs)~\cite{kenton2019bert, brown2020language, liu2019roberta, he2020deberta, he2021debertav3} play a crucial role in Natural Language Processing (NLP), offering substantial improvements in various downstream tasks~\cite{lee2020biobert, mars2022word, raffel2020exploring}. Customizing these models for specific tasks typically involves fine-tuning them to adapt pre-trained knowledge to particular requirements~\cite{alabi-etal-2022-adapting, uppaal-etal-2023-fine}. However, with the increasing scale of PLMs, the cost of full-model fine-tuning becomes prohibitive~\cite{qiu2020pre}. This has highlighted the demand for and increased interest in more parameter-efficient fine-tuning (PEFT) methods~\cite{zeng-etal-2023-one, ding2023parameter}.

Common PEFT methods introduce extra parameters to adapt downstream tasks and freeze all original parameters~\cite{li-liang-2021-prefix, liu-etal-2022-p, lester-etal-2021-power}. For instance, Low-Rank Adaptation (LoRA)~\cite{hu2022lora} has gained popularity for its streamlined approach by incorporating low-rank trainable matrices into existing fixed weight matrices in a PLM. However, LoRA assigns trainable parameters uniformly across all matrices, and there are studies~\cite{zhang2023adaptive} indicate that not all weights contribute equally to fine-tuning performance. This could result in inefficient parameter usage. Therefore, for optimal fine-tuning, is it possible to evaluate the parameter budget needs of each matrix and strategically allocate the limited parameters?

Fortunately, there are methods like AdaLoRA~\cite{zhang2023adaptive} that can alleviate the limitations of the prior PEFT methods by introducing a more nuanced parameter distribution strategy. Training with AdaLoRA begins with a higher parameter budget and simulates an SVD (Singular value decomposition) decomposition process, progressively pruning smaller singular values and corresponding singular vectors. It opens the door to implementing adaptive allocation of parameter budgets. However, its dependence on orthogonal regularization for the simulated SVD decomposition might restrict further improvements in fine-tuning efficiency. Additionally, the pruning strategy of AdaLoRA focuses solely on singular values and does not fully exploit all the available information in projection matrices, potentially leading to less-than-optimal decisions.

\begin{figure*}[ht]
  \centering
  \includegraphics[trim={0 50 0 50},clip,width=\textwidth]{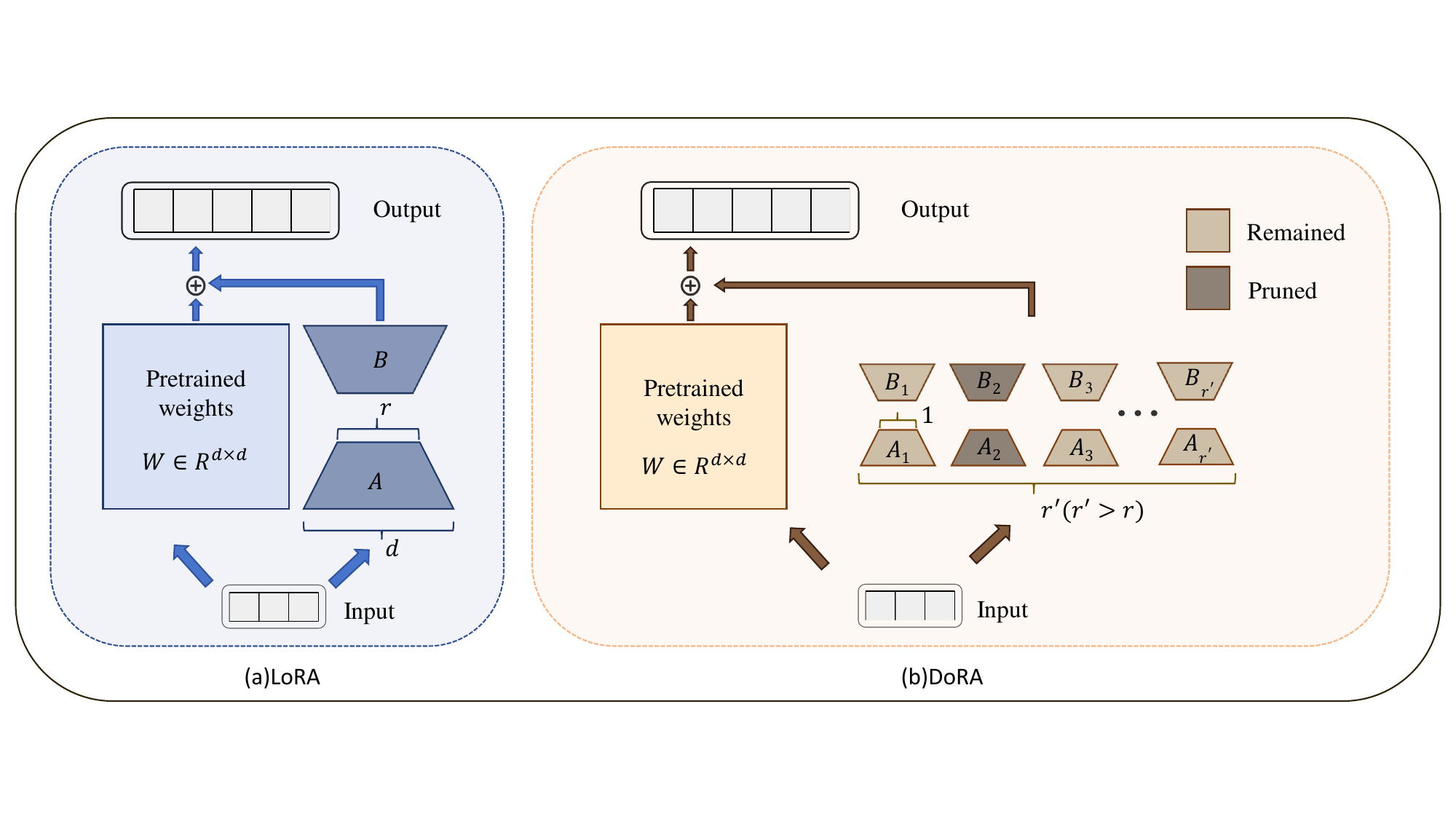}
  \caption{Figure (a) and Figure (b) illustrate the reparameterization of LoRA and DoRA. LoRA introduces a pair of low-rank matrices, A and B, each with a rank of $ r $, into the weight matrix. In contrast, DoRA introduces $ r' $ pairs of single-rank matrices, each acting as a LoRA component. During training, DoRA evaluates the contribution of each component to the overall performance and prunes components with smaller contributions, achieving adaptive allocation of parameters.}
  \label{fig:DoRA_compare_to_lora}
\end{figure*}

To address existing challenges, this work introduces the Dynamic Low-Rank Adaptation (DoRA) method, as depicted in Figure~\ref{fig:DoRA_compare_to_lora}. Different from LoRA approaches, DoRA decomposes high-rank LoRA layers into sums of single-rank components, evaluates the contribution of each component to overall performance, and prunes components with fewer contributions. This allows for the on-demand allocation of parameter budgets to modules of the PLM, maximizing the use of limited parameter budgets. Compared to existing methods of dynamic parameter allocation (e.g., AdaLoRA), DoRA can allocate parameter budgets more appropriately based on a richer set of information from projection matrices.

To sum up, our contributions are as follows:

\begin{itemize}
\item We introduce a novel PEFT method, DoRA, which surpasses the performance of full model fine-tuning with less than 0.3\% of the trainable parameters.
\item DoRA can efficiently identify the modules in PLMs that play a crucial role in the fine-tuning task, thereby allocating a larger parameter budget to these key modules.
\item DoRA maximizes the use of a limited parameter budget. Experiments demonstrate that DoRA outperforms baseline approaches across multiple downstream tasks under the same parameter budget constraints.
\end{itemize}

\begin{table*}[!t]
\centering
\resizebox{\textwidth}{!}{
\begin{tabular}{@{}cccc@{}}
\toprule
\makecell{Method} & \makecell{Parameter allocation strategy} & \makecell{Parametric method} & \makecell{Regularization penalty term} \\
\midrule
\makecell{LoRA} & \makecell{Equal allocation} & \makecell{Low-rank adaptation} & \makecell{N/A} \\
\makecell{AdaLoRA} & \makecell{Adaptive allocation} & \makecell{LoRA with SVD decomposition} & \makecell{SVD proximity} \\
\makecell{DoRA} & \makecell{Adaptive allocation} & \makecell{LoRA with component-wise decomposition} & \makecell{Component variance} \\
\bottomrule
\end{tabular}
}
\caption{Comparison of DoRA, LoRA, and AdaLoRA}
\label{table:compare_with_LoRA_AdaLoRA}
\end{table*}

\section{Background}
The emergence of PLMs such as BERT~\cite{kenton2019bert}, GPT~\cite{radford2019language}, and Llama~\cite{touvron2023llama} has meaningfully advanced the field of NLP. Trained on extensive text datasets, PLMs capture intricate linguistic patterns, enabling superior performance in various NLP tasks such as text categorization, named entity recognition, and machine translation~\cite{zhao2023survey}. Their flexibility in adapting to specific datasets via fine-tuning renders them exceedingly versatile for addressing various linguistic challenges.

PLMs predominantly leverage the Transformer architecture~\cite{vaswani2017attention} which features stacked Transformer blocks. Each block comprises two key components: a Multi-Head Attention (MHA) mechanism and a Feed-Forward Neural (FFN) network.
In particular, MHA effectively captures contextual relationships in text and is given by:
\begin{equation}
\label{eq:self_attention}
\begin{split}
\text{MHA}(x) = \text{Concatenate}(head_1(x),\\
head_2(x), \ldots, head_h(x))W_o
\end{split}
\end{equation}

\begin{equation}
\label{eq:MHA}
\begin{split}
head_i(x) = \text{Softmax}\left(\frac{(x W_{qi}) (x W_{ki})^T}{\sqrt{d_h}}\right)x W_{vi}
\end{split}
\end{equation}
where $ x \in \mathbb{R}^{n \times d} $ is the input feature, $ n $ is the sequence length and $ d $ is the hidden dimension. The mechanism consists $ h $ self-attention heads, each aiming to capture different aspects of information. For each head $ head_i $, there are three projection matrices: query $ W_{qi} $, key $ W_{ki} $, and value $ W_{vi} $, each with dimensions $ \mathbb{R}^{d \times d_h} $, where $ d_h $ is the dimension of each head, typically set to $ d/h $. The output projection matrix $ W_o \in \mathbb{R}^{d \times d} $ is used to produce the final output.

Attention scores are calculated by normalizing the dot product of queries and keys through the softmax function and are given by:
\begin{equation}
\label{eq:softmax}
\text{Softmax}(x_i) = \frac{e^{x_i}}{\sum_{j=1}^{n} e^{x_j}}
\end{equation}

These scores determine the attention each sequence position pays to other positions. Subsequently, these scores are multiplied by the value projection result to yield the output of each head. Finally, the outputs of all heads are concatenated and multiplied by the output projection matrix $ W_o $, forming the final MHA output.

Following MHA, the FFN further processes the information:
\begin{equation}
\label{eq:FFN}
\text{FFN}(x) = \text{ReLU}(xW_{f1} + b_1)W_{f2} + b_2
\end{equation}

This allows for more complex interactions between the features extracted by the self-attention mechanism. Each Transformer block incorporates a residual connection that adds the input of the block directly to its output. This approach helps to alleviate the vanishing gradient problem and ensures a consistent information flow across the layers of the models.

\begin{algorithm}[!t]
\caption{DoRA}
\label{algorithm:DoRA_algorithm}
\begin{algorithmic}[1]
\STATE \textbf{Input:} Dataset $D$; total steps $T$; initial budget $ b(0) $; final budget $ b(T) $; learning rate $ \gamma $; regularization coefficient $ \eta $; smoothing factor $ \beta $, DoRA parameters $ A $, $ B $, and $ c $.
\FOR{$t = 1$ to $T$}
    \STATE Sample a mini-batch $ d $ from $ D $ and compute the true label loss $ \mathcal{L}_{\text{true}} = \mathcal{L}(A, B, c, d) $;
    \STATE Compute the regularization loss $ \mathcal{L}_{\text{reg}} $ as Equation~\ref{eq:regularization_loss};
    \STATE Combine losses by adding true label loss and regularization loss $ \mathcal{L}_{\text{combined}} = \mathcal{L}_{\text{true}} + \eta \mathcal{L}_{\text{reg}} $;
    \STATE Perform backpropagation to compute the gradients of $ \mathcal{L}_{\text{combined}} $, and update the parameters with the learning rate $ \gamma $;
    \STATE Compute the importance score $ s $ as Equation~\ref{eq:importance_score}, update smoothed importance score $ \widetilde{s}(t) $ as Equation~\ref{eq:smoothed_score};
    \STATE Compute the current parameter budget $ b(t) $ as Equation~\ref{eq:budget_schedule};
    \STATE Prune the components with smaller $ \widetilde{s}(t) $ based on $ b(t) $, set their $ c $ to 0;
\ENDFOR
\STATE \textbf{Output:} The fine-tuned parameters $ \{A, B, c\} $.
\end{algorithmic}
\end{algorithm}

\section{Dynamic Low-Rank Adaptation}
\label{sec:Dynamic_Low_Rank_Adaptation}
In this paper, we aim to optimize the use of a limited parameter budget in fine-tuning PLMs with LoRA. We make improvements based on LoRA~\cite{hu2022lora} and AdaLoRA~\cite{zhang2023adaptive}, as shown in Table~\ref{table:compare_with_LoRA_AdaLoRA}. We propose DoRA that stands out due to its innovative approach, comprising three main strategies: a decomposition strategy that views a high-rank LoRA layer as a combination of multiple single-rank LoRA components, a dynamic rank allocation mechanism that adjusts these components based on their contribution to the overall performance of the model and a regularization penalty to ensure stable pruning throughout the process. The overall algorithm is shown in Algorithm~\ref{algorithm:DoRA_algorithm}.

\subsection{Parameterization}
DoRA introduces a novel perspective on PEFT for PLMs, building upon and enhancing the foundational LoRA technique. A standard LoRA layer is defined as:
\begin{equation}
\label{eq:lora_formula}
W = W_0 + \Delta W = W_0 + AB
\end{equation}
where $ W $ is the weight matrix after fine-tuning, $ W_0 $ denotes the original weight matrix, and $ A $, $ B $ are the low-rank matrices introduced by LoRA. By contrast, DoRA reinterprets this configuration and is given by:
\begin{equation}
\label{eq:DoRA_formula}
W = W_0 + \sum_{i=1}^{r'} \Delta W_i = W_0 + \sum_{i=1}^{r'}A_iB_ic_i
\end{equation}
here, $ r' $ represents the number of LoRA components, which will be explained in detail in Section~\ref{section:parameter_scheduling_and_pruning_strategy}. A LoRA component is a triplet of $ A_i $, $ B_i $, and $ c_i $, where $ A_i $ and $ B_i $ are single-rank matrices, shaped as $ d \times 1 $ and $ 1 \times d $ respectively. $ c_i $ is a scalar used for pruning the component, it is set to 0 if the component is to be pruned.

\subsection{Importance Scoring}
To evaluate the importance of each LoRA component, we employ an importance scoring mechanism that quantifies the contribution of each $ \Delta W_i $ and is given by:
\begin{equation}
\label{eq:importance_score}
\begin{split}
s_i = \| \Delta W_i \|_F / \| \sum_{j=1}^{r'} \Delta W_j \|_F \\
= \| A_iB_ic_i \|_F / \| \sum_{j=1}^{r'}A_jB_jc_j \|_F
\end{split}
\end{equation}
here, $ \|x\|_F $ denotes the Frobenius norm, a measure that calculates the square root of the sum of the squares of all elements in a matrix.

Employing the Frobenius norm allows us to measure the proportion of each LoRA component contribution to the total update magnitude of its corresponding LoRA layer. This metric facilitates an estimation of the potential impact on the total update of the LoRA layer if a particular component were to be pruned. Components with smaller impacts on the overall update magnitude are prioritized for pruning. This ensures that the pruning process minimally affects the performance, focusing on removing components that contribute the least to the effectiveness of the LoRA layer.

Compared to previous methods~\cite{zhang2023adaptive}, we use $ \|\Delta W_i\|_F $ instead of $ c_i $ to assess the importance of components, thereby incorporating information from $ A_i $ and $ B_i $ for a more comprehensive evaluation of component importance.

Moreover, to enhance the precision of the importance score, we employ a smoothing method by applying an exponential moving average to the importance scores. The smoothed importance score for the $ i $-th LoRA component at time $ t $, denoted as $ \widetilde{s}_i(t) $, blends the current importance score $ s_i $ with the previous one, adjusted by a factor $ \beta $:
\begin{equation}
\label{eq:smoothed_score}
\widetilde{s}_i(t) = \beta \cdot \widetilde{s}_i(t - 1) + (1 - \beta) \cdot s_i
\end{equation}

\subsection{Parameter Scheduling and Pruning Strategy}
\label{section:parameter_scheduling_and_pruning_strategy}
Parameter budget refers to the average number of LoRA components in each LoRA layer. It starts with an initial parameter budget,  $ b^{(0)} = r' $, which is deliberately set higher than the eventual target budget, $ b^{(T)} = r $, where $ r' $ and $ r $ are hyperparameters. Setting $ r' $ greater than $ r $ allows DoRA to explore a wider range of potential parameter allocations, facilitating the search for the optimal distribution. 

DoRA adopts a gentle pruning strategy. For the pruned triplets $ A_i $, $ B_i $, and $ c_i $, pruning is performed merely by setting $ c_i $ to $ 0 $ while keeping $ A_i $ and $ B_i $ unchanged. During subsequent training, the pruned triplets can be restored as long as $ c_i $ is updated to a non-zero value by backpropagation and is not pruned again.

DoRA warms up the training without pruning for the first $ t_i $ step, $ i $ denotes initial steps and then follows a cubic decrement pattern to prune components with lower importance scores until the remaining components reach the budget $ b^{(T)} $. Subsequently, it fixes the component distribution in the last $ t_f $ steps, $ f $ denotes the final steps. The overall budget scheduler is given by:
\begin{equation}
\label{eq:budget_schedule}
b^{(t)} =
\begin{cases}
b^{(0)} \qquad \text{if} ~ 0 \leq t < t_i,
\\ \\
\begin{aligned}
&b^{(0)} - \frac{(b^{(0)} - b^{(T)})}{b^{(0)}} \cdot \left(\frac{t - t_i}{t_f - t_i}\right)^3 \\
&\qquad \quad \, \, \, \, \text{if} ~ t_i \leq t \leq T - t_f,
\end{aligned}
\\ \\
b^{(T)} \qquad \text{if} ~ t > T - t_f.
\end{cases}
\end{equation}

\subsection{Dimensional Equilibrium Modulator}
DoRA utilizes the Frobenius norm of components for pruning, with a preference for clipping those with smaller norms. However, a potential issue arises when a component has most elements near zero and a few with considerably high values, leading to a relatively low Frobenius norm and thus being selected for pruning. This scenario can result in remarkable alterations in a limited number of dimensions of the total update, $ \Delta W $, resembling the effects of gradient explosion and adversely impacting model stability and fine-tuning performance.

To avoid this, we introduce the Dimensional Equilibrium Modulator (DEM) loss, which penalizes the variance of components as:
\begin{equation}
\label{eq:regularization_loss}
R = \frac{1}{n} \sum_{i=1}^{n} (\text{Var}(A_i) + \text{Var}(B_i)) 
\end{equation}
where $ \text{Var}(A_i) $ and $ \text{Var}(B_i) $ represent the variances of components $ A_i $ and $ B_i $, with $ n $ indicating the number of components. DEM encourages a uniform distribution of elements within components, avoiding disproportionate impacts from isolated or few dimensions, effectively reducing perturbations from model pruning, and enhancing model stability.

\section{Experiments}

\subsection{Experimental Setup}
We compared DoRA with existing baseline methods to evaluate its performance in natural language understanding (NLU), question answering (QA), and text generation (summarization) tasks. We chose RoBERTa~\cite{liu2019roberta} and Bart~\cite{Mike2019bart} as the foundational models, used respectively for NLU and QA tasks, and for summarization tasks.

RoBERTa is an optimized version of the BERT~\cite{kenton2019bert} architecture, which significantly improves performance on a variety of language understanding tasks through extended training, larger datasets, and finer tuning of parameters. Bart is a Transformer-based~\cite{vaswani2017attention} sequence-to-sequence pre-trained model specifically designed for text generation tasks, such as summarization. It effectively handles various generation tasks by combining bidirectional and autoregressive Transformer architectures.

We tested the performance on several standard datasets: using the GLUE (General Language Understanding Evaluation)~\cite{wang-etal-2018-glue} dataset to evaluate NLU tasks, SQuAD~\cite{rajpurkar-etal-2016-squad} (Stanford Question Answering Dataset) for QA, and Xsum~\cite{Narayan2018DontGM} for text summarization. GLUE is a set of dataset for training and testing NLU systems, including various tasks such as sentiment analysis and textual entailment. SQuAD is a question answering dataset that consists of questions generated from Wikipedia articles and their corresponding answers. Xsum provides a testing environment for extreme summarization tasks aimed at generating single-sentence summaries, challenging the models under extreme information compression conditions.

We selected several mainstream fine-tuning methods as baselines, including LoRA, AdaLoRA, Adapter Tuning, BitFit, and full model fine-tuning. LoRA fine-tunes the model weights by adding low-rank matrices to the pre-trained matrices; AdaLoRA is an improvement of LoRA, adding an adaptive adjustment mechanism. Adapter Tuning fine-tunes by inserting lightweight network modules into the PLM. BitFit adjusts only the bias parameters in the PLM. Full model fine-tuning is a traditional method that involves comprehensive adjustment of all model weights.

We report the average results based on 5 random seeds, as shown in Table~\ref{table:glue_result}, Table~\ref{table:squad_result}, and Table~\ref{table:xsum_result}, The hyperparameter settings for the experiments can be found in Appendix~\ref{section:hyper_parameter}.

\begin{table*}[!t]
\centering
\small
\begin{tabular}{@{}c|c|cccccccc|c@{}}
\toprule
\makecell{Method} & \makecell{Trainable \\ Parameters} &
\makecell{RTE}     & \makecell{MRPC}    & \makecell{STS-B} & \makecell{CoLA}  & \makecell{SST-2} & \makecell{QNLI}  & \makecell{QQP}   & \makecell{MNLI}  & \makecell{Avg} \\
\midrule
\makecell{Full FT} & \makecell{124.65M} &
\makecell{78.63}   & \makecell{88.33}   & \makecell{90.31} & \makecell{60.26} & \makecell{94.73} & \makecell{92.58} & \makecell{90.75} & \makecell{87.68} & \makecell{85.41} \\
\makecell{BitFit}  & \makecell{0.10M}   &
\makecell{79.57}   & \makecell{89.07}   & \makecell{90.55} & \makecell{61.16} & \makecell{94.38} & \makecell{90.99} & \makecell{88.08} & \makecell{85.50} & \makecell{84.91} \\
\hline
\makecell{H-Adapter} & \makecell{1.20M} &
\makecell{80.43}     & \makecell{89.90} & \makecell{90.16} & \makecell{62.62} & \makecell{93.73} & \makecell{92.82} & \makecell{90.83} & \makecell{86.53} & \makecell{85.88} \\
\makecell{P-Adapter} & \makecell{1.19M} &
\makecell{80.51}     & \makecell{89.51} & \makecell{90.65} & \makecell{63.87} & \makecell{93.83} & \makecell{92.61} & \makecell{90.53} & \makecell{86.75} & \makecell{86.03} \\
\makecell{LoRA}      & \makecell{1.33M} &
\makecell{80.65}     & \makecell{89.90} & \makecell{90.91} & \makecell{63.54} & \makecell{93.71} & \makecell{92.76} & \makecell{90.44} & \makecell{87.11} & \makecell{86.13} \\
\makecell{AdaLoRA}   & \makecell{1.27M} &
\makecell{81.23}     & \makecell{89.02} & \makecell{91.22} & \makecell{63.23} & \makecell{95.11} & \makecell{92.84} & \makecell{90.48} & \makecell{\textbf{87.89}} & \makecell{86.38} \\
\makecell{DoRA}      & \makecell{1.31M} &
\makecell{\textbf{81.73}} & \makecell{\textbf{90.05}} & \makecell{\textbf{91.34}} & \makecell{\textbf{65.35}} & \makecell{\textbf{95.21}} & \makecell{\textbf{92.97}} & \makecell{\textbf{91.32}} & \makecell{87.81} & \makecell{\textbf{86.97}} \\
\hline
\makecell{H-Adapter} & \makecell{0.31M} &
\makecell{78.56}     & \makecell{88.64} & \makecell{90.88} & \makecell{61.76} & \makecell{93.54} & \makecell{92.52} & \makecell{90.16} & \makecell{86.31} & \makecell{85.30} \\
\makecell{P-Adapter} & \makecell{0.30M} &
\makecell{79.07}     & \makecell{88.74} & \makecell{90.44} & \makecell{62.92} & \makecell{93.24} & \makecell{92.59} & \makecell{89.94} & \makecell{86.23} & \makecell{85.40} \\
\makecell{LoRA}      & \makecell{0.33M} &
\makecell{78.63}     & \makecell{88.68} & \makecell{91.24} & \makecell{62.40} & \makecell{93.25} & \makecell{92.75} & \makecell{90.12} & \makecell{87.01} & \makecell{85.50} \\
\makecell{AdaLoRA}   & \makecell{0.32M} &
\makecell{79.04}     & \makecell{88.81} & \makecell{91.06} & \makecell{63.17} & \makecell{94.79} & \makecell{92.87} & \makecell{90.07} & \makecell{\textbf{87.64}} & \makecell{85.93} \\
\makecell{DoRA}      & \makecell{0.34M} &
\makecell{\textbf{79.15}} & \makecell{\textbf{89.72}} & \makecell{\textbf{91.28}} & \makecell{\textbf{64.90}} & \makecell{\textbf{94.98}} & \makecell{\textbf{92.93}} & \makecell{\textbf{90.64}} & \makecell{87.45} & \makecell{\textbf{86.38}} \\
\bottomrule
\end{tabular}
\caption{Results of fine-tuning RoBERTa base on GLUE. We report results on development set, Pearson correlation for STS-B, Matthew's correlation for CoLA, average accuracy for MNLI (matched and mismatched), and accuracy for other tasks. ``Full FT'', ``H-Adapter'', and ``P-Adapter'' represent full fine-tuning, Houlsby adapter, and Pfeiffer adapter. The best results are in \textbf{bold}.}
\label{table:glue_result}
\end{table*}

\begin{table}
\centering
\resizebox{\columnwidth}{!}{
\begin{tabular}{cccc}
\toprule
   & Trainable Parameters & SQuAD v1 & SQuAD v2 \\
\midrule
Full FT   & 124.65M & 85.32/91.49          & 79.95/83.09\\
BitFit    & 0.10M   & 82.34/89.45          & 74.28/77.46\\
\hline
H-Adapter & 1.20M   & 84.95/91.07          & 79.14/82.08\\
P-Adapter & 1.19M   & 84.86/90.86          & 78.86/81.84\\
LoRA      & 1.33M   & 85.13/91.39          & 79.25/82.34\\
AdaLoRA   & 1.28M   & 85.34/91.72          & 79.87/82.84\\
DoRA      & 1.30M   & \textbf{85.97/92.24} & \textbf{80.43/83.53}\\
\hline
H-Adapter & 0.31M   & 84.60/90.44          & 78.48/81.55\\
P-Adapter & 0.30M   & 84.44/90.34          & 78.22/81.34\\
LoRA      & 0.33M   & 84.91/90.91          & 78.83/81.78\\
AdaLoRA   & 0.32M   & 85.13/91.32          & 79.47/82.40 \\
DoRA      & 0.34M   & \textbf{85.73/91.88} & \textbf{79.90/82.92} \\
\bottomrule
\end{tabular}
}
\caption{Results of fine-tuning RoBERTa based on SQuAD. We report the exact match and F1 scores on the development set. The best results are in \textbf{bold}.}
\label{table:squad_result}
\end{table}

\begin{table}
\centering
\resizebox{\columnwidth}{!}{
\begin{tabular}{ccc}
\toprule
   & Trainable Parameters & Xsum\\
\midrule
Full FT & 124.65M & 40.61/17.76/32.91\\
\hline
LoRA    & 1.33M   & 38.77/15.63/30.66\\
AdaLoRA & 1.27M   & 39.14/16.23/31.34\\
DoRA    & 1.31M   & \textbf{39.67/16.73/31.78}\\
\hline
LoRA    & 0.33M   & 37.17/14.57/29.72\\
AdaLoRA & 0.32M   & 38.32/15.69/30.74\\
DoRA    & 0.34M   & \textbf{38.94/16.22/31.36}\\
\bottomrule
\end{tabular}
}
\caption{Results of fine-tuning Bart base on Xsum. We report the Rouge-1, Rouge-2, and Rouge-L scores on the development set. The best results are in \textbf{bold}.}
\label{table:xsum_result}
\end{table}

\subsection{Results}
We investigate the performance of DoRA and baseline methods across subtasks of the GLUE benchmark, conducting experiments under two different parameter budget scenarios. 

As shown in Table~\ref{table:glue_result}, DoRA and AdaLoRA, employing adaptive parameter allocation strategies, outperform all baseline methods using uniform parameter distribution, demonstrating the remarkable effectiveness of adaptive parameter allocation. Across the GLUE benchmark, DoRA surpasses LoRA by 0.84\% and 0.88\%, and AdaLoRA by 0.59\% and 0.45\% under two parameter budgets, further proving the broad applicability and effectiveness of DoRA's adaptive parameter allocation strategy in multiple tasks.

Especially noteworthy is DoRA's performance on the CoLA dataset, where it shows the highest improvement, surpassing the highest performing baseline method by 1.48\% and 1.73\% under two parameter budgets. This highlights DoRA's advantage in handling the linguistic acceptability task, showcasing its efficiency in dealing with challenging NLP tasks. However, DoRA's performance on the MNLI task slightly lags behind AdaLoRA, likely due to MNLI being the largest dataset in GLUE with high task complexity, indicating a need for further optimization of the adaptive parameter allocation strategy when dealing with large-scale complex tasks.

It is worth mentioning that DoRA demonstrates exceptional parameter efficiency, surpassing the performance of full model fine-tuning with only 0.34M parameters, less than 0.3\% of full model fine-tuning, highlighting DoRA's capability in effectively utilizing a limited parameter budget.

Similar results are also observed in the experiments on SQuAD and Xsum, where DoRA outperformes all baseline PEFT methods under both parameter settings.

\section{Analysis and Discussion}

\subsection{Effectiveness of DEM}
To verify the effectiveness of DEM, we tested fine-tuning on datasets including STS-B, CoLA, and SST-2, with and without DEM, without DEM means setting hyper-parameter regularization coefficient $ \eta $ to 0, as shown in Table~\ref{table:Performance_comparison_of_DEM}. 

\begin{table}[ht]
\centering
\small
\begin{tabular}{@{}cccc@{}}
\toprule
Model & STS-B & CoLA & SST-2 \\
\midrule
\textit{with DEM}    & \textbf{91.34} & \textbf{65.35} & \textbf{95.21} \\
\textit{without DEM} & 91.23          & 64.17          & 95.12 \\
\bottomrule
\end{tabular}
\caption{Performance comparison of DEM}
\label{table:Performance_comparison_of_DEM}
\end{table}

Enabling DEM imposes penalties on the variance of LoRA components, encouraging a uniform weight distribution, and avoiding extreme variations in overall update $ \Delta W $ across a few dimensions due to pruning. Fine-tuning with DEM enabled achieved higher results, demonstrating the effectiveness of DEM.
 
\subsection{Parameter Allocation Preference}

\begin{figure*}[ht]
\centering
\includegraphics[width=\textwidth]{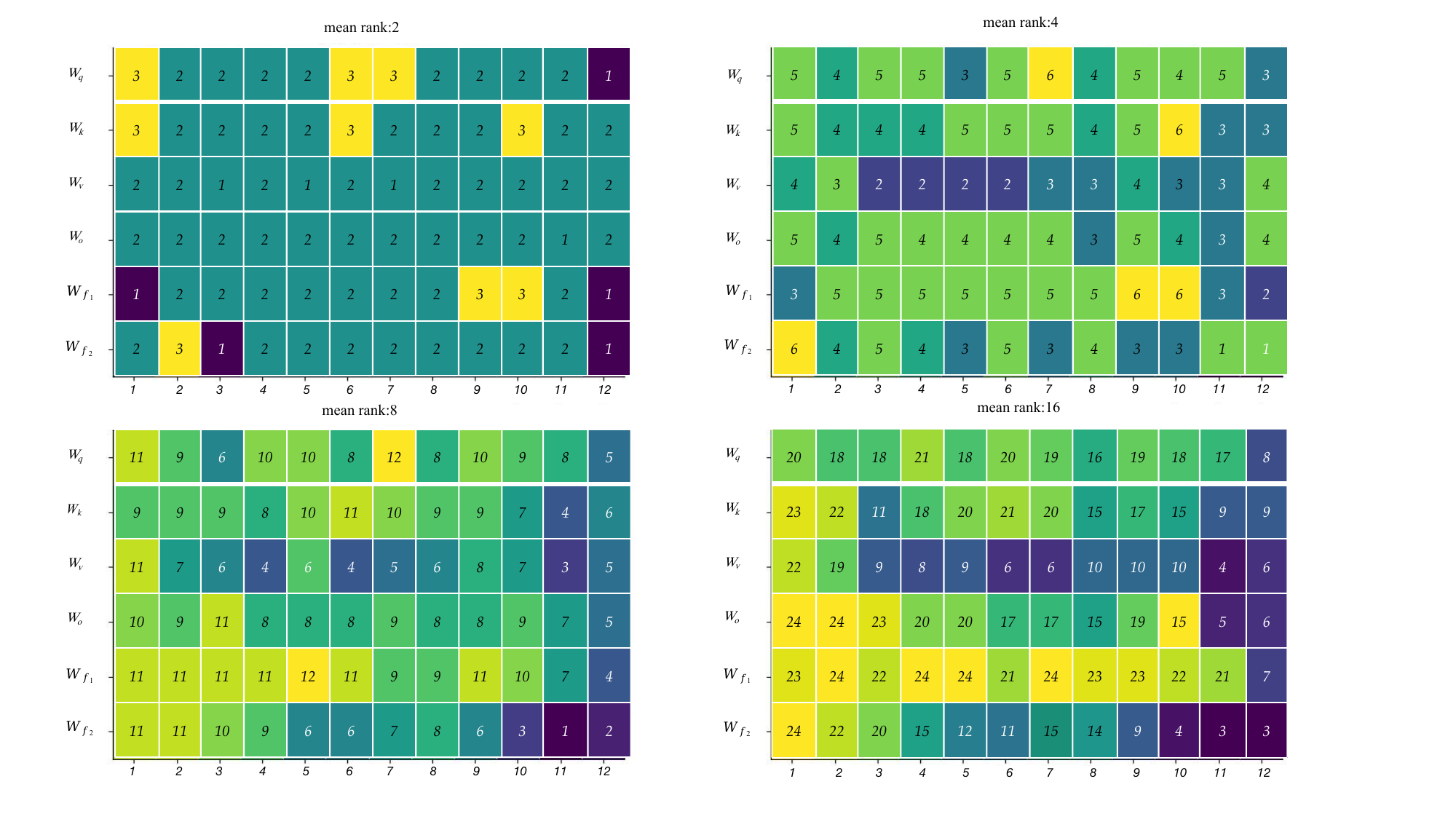}
\caption{Rank distribution under four parameter budgets}
\label{fig:Parameter_Allocation_Preference}
\end{figure*}

To validate whether DoRA can identify key modules in PLMs, we set the final budgets $ b^{(T)} $ to 2, 4, 8, and 16, with 1.5 times the final budget as the initial budget $ b^{(0)} $, and conducted fine-tuning experiments on SST-2 dataset respectively.

The results are visually presented in Figure~\ref{fig:Parameter_Allocation_Preference}, which shows that, in the intermediate layers, the query and key matrices are allocated with more parameter budget, while the value matrices are allocated with fewer budget. The initial output matrices receive more budget. In the feed-forward neural network, the lower projection matrices, represented as $ W_{f_{2}} $ in the figure, at the backend, especially in the last few layers, are allocated with very few budgets.

DoRA exhibited the same parameter allocation tendencies across all four configurations, demonstrating its ability to consistently identify key modules in PLMs and allocate more parameter budgets to them accordingly.

\subsection{Impact of Initial Budget}
We investigated the impact of the initial budget $ b^{(0)} $  across the MRPC, STS-B, and SST-2 datasets. We fine-tuned models starting from various initial budgets and pruned them to a consistent final budget $ b^{(T)} $ of 2. The results are presented in Table~\ref{table:The_Impact_of_Initial_Rank_Size}. The first row indicates that when the initial budget is 2, it matches the final budget, which means no model pruning was performed. 

Intriguingly, our findings suggest that maintaining a constant final parameter budget while starting with a higher initial parameter budget improves model performance. We attribute this improvement to a more generous initial parameter budget offering a wider exploration space for DoRA, thereby increasing the chance of preserving essential parameters during pruning and optimizing the model's final performance.

\begin{table}[ht]
\centering
\small
\begin{tabular}{@{}cccc@{}}
\toprule
Initial Budget& MRPC & STS-B & SST-2 \\
\midrule
\textit{2} & 76.32 & 90.97 & 94.88 \\
\textit{3} & 89.74 & 91.28 & 94.98 \\
\textit{4} & 89.92 & 91.40 & 95.01 \\
\textit{6} & 90.14 & 91.62 & 95.05 \\
\textit{8} & 90.17 & 91.65 & 95.11 \\
\bottomrule
\end{tabular}
\caption{The Impact of Initial Rank Size}
\label{table:The_Impact_of_Initial_Rank_Size}
\end{table}

\section{Related Work}
PEFT is crucial for the fine-tuning of PLMs in practical applications. These techniques primarily focus on updating a select subset of the model's parameters or introducing new parameters on a small scale, enabling more efficient use of resources. These approaches are particularly valuable in scenarios constrained by computational resources. Existing PEFT methods can generally be divided into three categories: addition-based methods, specification-based methods, and reparameterization-based methods~\cite{ding2022delta}.\\

\noindent \textbf{Addition-based methods} achieve adjustment by adding extra modules or trainable parameters to PLMs, such as trainable adapters or soft prompts. These methods are scalable and applicable to models of various sizes, with the performance gap between them and full model fine-tuning narrows as model size increases. Examples include adapters~\cite{pmlr-v97-houlsby19a, pfeiffer-etal-2021-adapterfusion, he2021towards, zhu2021counter}, which insert small neural modules into transformer layers for adjustment, and prompt-based tuning~\cite{li2021prefix, gao-etal-2021-making, hu-etal-2022-knowledgeable, tan-etal-2022-msp, lester2021power, vu2021spot}, which stimulates PLMs by adding additional context around the original input.\\

\noindent \textbf{Specification-based methods} focus on fine-tuning a few inherent parameters within the model without altering its internal structure~\cite{vucetic2022efficient, holmes2021nxmtransformer}. By directly specifying which parts of the parameters to optimize, these approaches achieve efficient model adaptation, maintaining performance close to full parameter fine-tuning while reducing the number of parameters adjusted. Examples include BitFit~\cite{ben-zaken-etal-2022-bitfit}, which optimizes only the bias terms in the model, and Diff Pruning~\cite{guo-etal-2021-parameter}, which introduces sparsity by optimizing a difference vector.\\

\noindent \textbf{Reparameterization-based methods} optimize models by transforming adaptive parameters into more efficient forms, often based on the low-rank hypothesis. These methods~\cite{holmes2021nxmtransformer, karimi2021compacter, edalati2022krona, zhang2023adaptive, lialin2023relora, ding2023sparse, valipour2023dylora, su2024mixture, liu2024dora} aim to reduce computational and memory costs by optimizing low-dimensional proxy parameters while maintaining or surpassing the performance of full parameter fine-tuning. They are grounded in the theory that PLM adaptations to downstream tasks are inherently low-rank. Examples include LoRA~\cite{hu2022lora}, which optimizes based on the hypothesis of a low intrinsic rank of weight changes.

\section{Conclusion}
In this paper, we introduce Dynamic Low-Rank Adaptation (DoRA), a novel method aiming at enhancing the efficiency of fine-tuning PLMs by dynamically adjusting parameter distribution. DoRA innovatively allocates parameter budgets based on their importance to specific tasks, demonstrating considerable improvements in NLP applications. Experimental results indicate that DoRA surpasses baseline methods, highlighting its potential for broader adoption in model optimization efforts.

The innovation of DoRA lies in its adoption of an adaptive parameter allocation strategy, which, unlike traditional uniform distribution, dynamically adjusts the distribution of parameter budgets based on their contribution. Additionally, DoRA employs a component-wise decomposition approach for handling LoRA layers, treating high-rank LoRA layers as a combination of single-rank LoRA components. These components are adjusted through a dynamic rank allocation mechanism, pruned according to their contribution to the overall model performance. To ensure stable pruning throughout the process, DoRA incorporates a regularization penalty term focused on reducing component variance.

\section*{Limitation}
Our study confirms the effectiveness of DoRA in several NLP tasks. However, its evaluation has been limited to these tasks, and its efficacy in handling more complex NLP challenges, such as machine translation or multimodal tasks, has yet to be established. Moreover, the models used in our experiments are somewhat limited in scale, as we have not conducted experiments with large language models (LLMs). Addressing this limitation, future work could explore DoRA's potential in these sophisticated areas of NLP.

\section*{Acknowledgements}
This work is supported by the National Natural Science Foundation of China (No.62376019, 61976015, 61976016, 61876198 and 61370130) and the Talent Fund of Beijing Jiaotong University (2024JBRC005). We sincerely thank the reviewers for their insightful comments and suggestions to improve the quality of the paper.

\bibliography{anthology, custom}

\clearpage
\appendix
\section{Potential Risks of Our Method}
Since our proposed method adapts pre-trained models to specific tasks, it has the potential to extend the range of languages supported by these models. However, there is a risk that some malicious users might exploit this new capability to provide services in politically sensitive languages and tasks. For instance, a malicious user could use our method to generate hateful or offensive sentences in some politically sensitive languages.

\section{Dataset Detail}
\textbf{GLUE}~\cite{wang-etal-2018-glue} benchmark is a collection of diverse natural language understanding tasks designed to evaluate and analyze the performance of models across a wide range of linguistic challenges. The benchmark encompasses a variety of tasks, including linguistic acceptability (CoLA~\cite{warstadt-etal-2019-neural}), sentiment analysis (SST-2~\cite{socher-etal-2013-recursive}), paraphrase detection (MRPC~\cite{dolan-brockett-2005-automatically}, QQP~\cite{wang-etal-2018-glue}), semantic textual similarity (STS-B~\cite{wang-etal-2018-glue}), and natural language inference (MNLI~\cite{williams-etal-2018-broad}, QNLI~\cite{rajpurkar-etal-2016-squad}, RTE~\cite{bentivogli2009fifth}). The dataset statistics are presented in Table~\ref{table:GLUE_Statistics}.

\begin{table}[!ht]
\centering
\small
\begin{tabular}{@{}lcccc@{}}
\toprule
Corpus  & Train   & Valid   & Test   & Metrics \\
\midrule
RTE     & 2.5k    & 277     & 3k     & Accuracy \\
MRPC    & 3.7k    & 408     & 1.7k   & Accuracy\\
STS-B   & 5.7k    & 1.5k    & 1.4k   & Pearson corr\\
CoLA    & 8.5k    & 1,043   & 1,063  & Matthews corr \\
SST-2   & 67k     & 872     & 1.8k   & Accuracy \\
QNLI    & 105k    & 5.5k    & 5.5k   & Accuracy \\
QQP     & 364k    & 40.4k   & 391k   & Accuracy\\
MNLI    & 393k    & 20k     & 20k    & Accuracy \\
\bottomrule
\end{tabular}
\caption{Statistics of the GLUE Benchmark Datasets}
\label{table:GLUE_Statistics}
\end{table}

\textbf{SQuAD}~\cite{rajpurkar-etal-2016-squad} is an extensive reading comprehension dataset aimed at evaluating the ability of models to understand and answer questions based on Wikipedia article contents. The dataset features two major versions: SQuAD 1.1 and SQuAD 2.0. SQuAD 1.1 consists of over 100,000 question-answer pairs on 500+ articles, where questions are posed by crowd workers on a given passage and the answers are segments of text from the passage. SQuAD 2.0 extends the SQuAD 1.1 dataset with over 50,000 additional unanswerable questions that are written adversarially by crowd workers to look similar to answerable ones but do not have answers in the text. This makes SQuAD 2.0 more challenging and helps models better emulate human reading comprehension abilities by not only retrieving answers but also determining when no valid answer exists within the text.

\textbf{Xsum}~\cite{Narayan2018DontGM} is designed specifically for the task of single-sentence news summarization to create a concise abstract of a news article. Xsum consists of approximately 227,000 articles collected from the British Broadcasting Corporation (BBC). Each article comes with a professionally written, single-sentence summary, making it particularly challenging for models due to the extreme brevity and the need for models to abstract rather than simply extract content. Unlike other summarization datasets, which often focus on extracting salient sentences directly from the text, Xsum requires models to generate informative, concise, and grammatically coherent summaries that capture the core essence of the article content, often requiring synthesis and rephrasing skills beyond mere extraction.

\section{Baseline Detail}
\textbf{LoRA}~\cite{hu2022lora}: LoRA achieves fine-tuning by integrating low-rank matrices within the weight matrices of the PLM. It maintains efficiency by adjusting a reduced set of parameters, which is particularly advantageous for scaling to large models.

\textbf{AdaLoRA}~\cite{zhang2023adaptive}: Building upon LoRA, AdaLoRA optimizes parameter utilization by adaptively adjusting the parameter budget throughout the training process. This adaptive strategy improves fine-tuning efficiency and has demonstrated enhanced performance in a variety of NLP tasks.

\textbf{Adapter Tuning}~\cite{pmlr-v97-houlsby19a, rebuffi2017learning, pfeiffer-etal-2021-adapterfusion}: Incorporates methods such as the Houlsby Adapter (H-Adapter) and Pfeiffer Adapter (P-Adapter). Adapter Tuning fine-tunes models by inserting small, trainable adapter modules between existing layers of the PLM. This approach does not modify the original weights in the PLM, offering a flexible yet conservative way to adapt the model to new tasks.

\textbf{BitFit}~\cite{ben-zaken-etal-2022-bitfit}: An minimalistic PEFT method that focuses solely on adjusting the bias terms within the model. BitFit has been shown to achieve performance levels comparable to those of full model fine-tuning under specific conditions.

\textbf{Full Model Fine-tuning}: This conventional method updates all the parameters of a PLM to tailor it to specific downstream tasks. Full model fine-tuning demands remarkable computational resources.

\section{Training Setup}
Our experiments were conducted using the PyTorch~\cite{paszke2019pytorch} framework, with models and datasets sourced from the Huggingface~\cite{wolf-etal-2020-transformers} platform. The computations were performed on NVIDIA GeForce RTX 3090 GPUs.

We refer to LoRA's initialization method, applying Kaiming initialization~\cite{He_2015_ICCV} to $ A_i $ and $ B_i $, and initializing $ c_i $ with 0. This ensures that the initial update amount $ \Delta W_i $ of each component is 0, preserving the behavior of the original model at the initial stage of training.

\section{Hyper Parameter}
\label{section:hyper_parameter}
We fix the following hyperparameters across all experiments:

\begin{itemize}
\item Initial budget $ b^{(0)} $: 1.5 times the final budget $ b^{(T)} $
\item Smoothing factor $ \beta $: 0.9
\item Initial steps $ t_i $: 15\% of the total steps
\item Final steps $ t_f $: 50\% of the total steps
\end{itemize}

We fixed sentence length and searched remain hyperparameters, including the learning rate, batch size, number of training epochs, regularization coefficient, and pruning step interval, as shown in Table~\ref{table:Hyper_Parameter}
\begin{table*}[!t]
\centering
\begin{tabular}{@{}llccccc@{}}
\toprule
Corpus   &length& learning rate & batch size & epochs & regularization coefficient & pruning step interval \\
\midrule
RTE      & 128 & 2e-03   & 16 & 80 & 0.5 & 10 \\
MRPC     & 128 & 2.5e-03 & 16 & 30 & 0.3 & 50 \\
STS-B    & 128 & 3e-03   & 16 & 45 & 0.3 & 10 \\
CoLA     & 128 & 2e-03   & 16 & 45 & 0.3 & 10 \\
SST-2    & 128 & 8e-04   & 64 & 60 & 0.5 & 10 \\
QNLI     & 128 & 3.5e-03 & 16 & 12 & 0.5 & 50 \\
QQP      & 128 & 1e-03   & 16 & 10 & 0.3 & 50 \\
MNLI     & 128 & 3e-03   & 32 & 10 & 0.3 & 50 \\
SQuAD v1 & 384 & 5e-03   & 16 & 14 & 0.1 & 100\\
SQuAD v2 & 384 & 3.4e-03 & 16 & 14 & 0.1 & 100\\
Xsum     & 768 & 3.4e-03 & 16 & 15 & 0.1 & 100\\
\bottomrule
\end{tabular}
\caption{Hyper-parameter setup}
\label{table:Hyper_Parameter}
\end{table*}

\section{Computing Efficiency}
We evaluate the computational efficiency of DoRA compared to baseline methods based on the training time per epoch on QNLI dataset and GPU memory usage. The results are shown in Table~\ref{tab:computing_efficiency}. DoRA's computational efficiency is comparable to AdaLoRA, slightly lower than LoRA, and significantly higher than full model fine-tuning. 

\begin{table*}
\centering
\begin{tabular}{ccc}
\toprule
&  Time Per Epoch/min& GPU memory consumption/MiB\\
\midrule
DoRA      & 7.8  & 4112\\
AdaLoRA   & 7.9  & 4130\\
LoRA      & 7.2  & 4094\\
H-Adapter & 6.8  & 3698\\
P-Adapter & 6.4  & 3678\\
BitFit    & 6.4  & 3612\\
Full FT   & 10.3 & 6104\\
\bottomrule
\end{tabular}
\caption{Computing Efficiency Comparison}
\label{tab:computing_efficiency}
\end{table*}

\end{document}